\documentclass{article}
\usepackage{spconf,amsmath,graphicx}

\usepackage{amssymb} % define this before the line numbering.
\usepackage{color}
\usepackage{url}
\usepackage[tight,footnotesize]{subfigure}
\usepackage{multirow}

\usepackage{gensymb}

\title{UltraCompression: Framework for High Density Compression of Ultrasound Volumes using Physics Modeling Deep Neural Networks}
\usepackage{spconf}
\makeatletter
\def\@name{ \emph{Debarghya China$^{\star \dagger}$\thanks{$^{\star}$Equally contributing authors.}, Francis Tom$^{\star \dagger}$, Sumanth Nandamuri$^{\star \dagger}$, Aupendu Kar$^{\dagger}$},  \\ \emph{Mukundhan Srinivasan$^{\mp}$, Pabitra Mitra$^{\dagger}$, Debdoot Sheet$^{\dagger}$}}
\makeatother

\address{$^{\dagger}$Indian Institute of Technology Kharagpur, Kharagpur, India \\ $^{\mp}$NVIDIA, Bengaluru, India}

\begin{document}
%\ninept
%
\maketitle
\begin{abstract}

Ultrasound image compression by preserving speckle-based key information is a  challenging task. In this paper, we introduce an ultrasound image compression framework with the ability to retain realism of speckle appearance despite achieving very high-density compression factors. The compressor employs a tissue segmentation method, transmitting segments along with transducer frequency, number of samples and image size as essential information required for decompression. The decompressor is based on a convolutional network trained to generate patho-realistic ultrasound images which convey essential information pertinent to tissue pathology visible in the images. We demonstrate generalizability of the building blocks using two variants to build the compressor. We have evaluated the quality of decompressed images using distortion losses as well as perception loss and compared it with other off the shelf solutions. The proposed method achieves a compression ratio of $725:1$ while preserving the statistical distribution of speckles. This enables image segmentation on decompressed images to achieve dice score of $0.89 \pm 0.11$, which evidently is not so accurately achievable when images are compressed with current standards like JPEG, JPEG 2000, WebP and BPG. We envision this frame work to serve as a roadmap for speckle image compression standards.

\end{abstract}

\begin{keywords}
Image compression, ultrasound imaging, generative models, adversarial learning
\end{keywords}

\begin{figure}[t!]
\centering
\subfigure[Framework]{\label{framework} {\setlength\fboxsep{0pt}\setlength\fboxrule{0pt}\fbox{\includegraphics[width=0.4\textwidth]{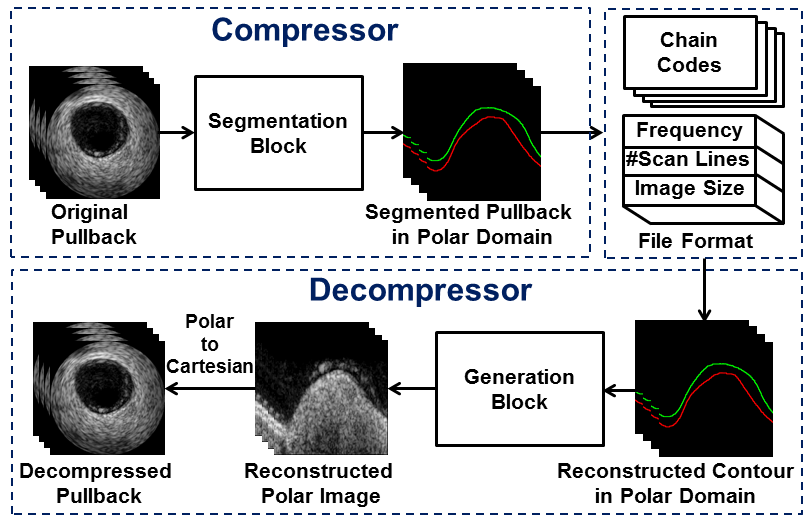}}}}
\subfigure[Original Image]{\label{org1} {\setlength\fboxsep{0pt}\setlength\fboxrule{0pt}\fbox{\includegraphics[width=0.14\textwidth]{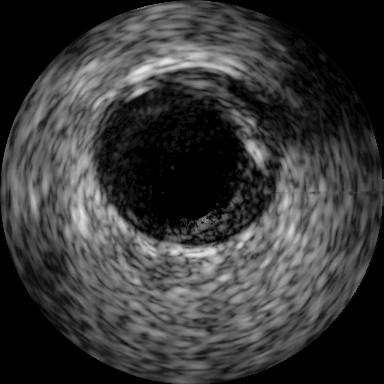}}}}
  \subfigure[UltraCompression]{\label{Ultracompression1} {\setlength\fboxsep{0pt}\setlength\fboxrule{0pt}\fbox{\includegraphics[width=0.14\textwidth]{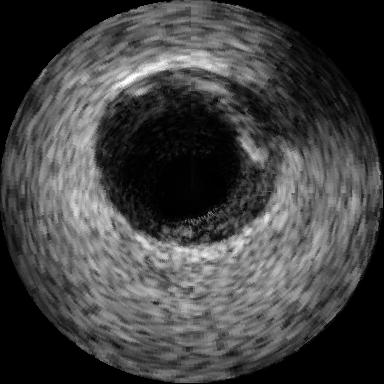}}}}
 \subfigure[PD var]{\label{Perception} {\setlength\fboxsep{0pt}\setlength\fboxrule{0pt}\fbox{\includegraphics[width=0.15\textwidth]{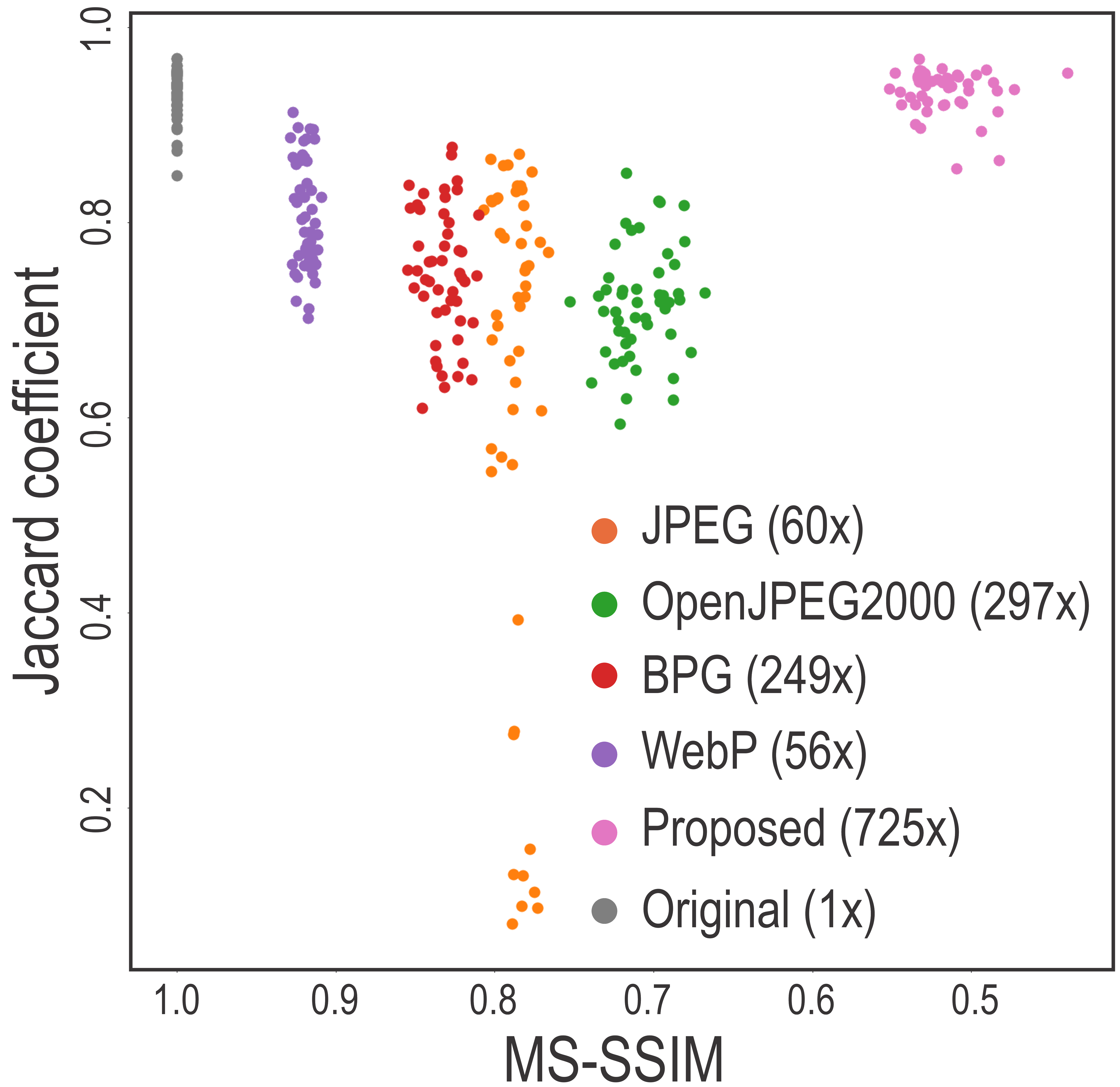}}}}
\caption{(a) Frame work, (b) original image (c) decompressed output and (d) perception distortion (PD) variance comparison with existing compression standards.}
\label{fig:UltraCompression}
\end{figure}

\section{Introduction}

Ultrasound (US) has been in use for medical imaging for more than three decades, on account of its salient advantages which include the relatively low cost of ownership and operation, non-ionizing nature of radiation, real-time imaging capability, high resolution, and ability to serve for both inside-out and outside-in imaging~\cite{china2018anatomical}. In recent years there has been significant advancements beyond the conventional 2D US imaging techniques with the inclusion of 2D+T, 3D, 3D+T modes while making them more advanced and increasing their deployment in mainline medical imaging~\cite{liu2017current}. This has resulted in an increase in the quantity of data being used to store these images, thereby creating difficulties in storage and transmission. Even though image compression techniques have been able to solve this problem in general for consumer grade camera images and some medical images, ultrasound image compression is challenging due to the presence of speckle patterns which are hard to preserve when compressed with existing standards like JPEG, JPEG2000 \cite{liu2017current}. In this paper, we propose a framework for ultrasound image compression wherein the compressor and the decompressor are designed taking into account the physics of ultrasound image formation process as illustrated in Fig.~\ref{framework}, to obtain decompressed images (Fig.~\ref{Ultracompression1}) which are perceivably similar to the original image (Fig.~\ref{org1}), as also evident in Fig.~\ref{Perception} representing the perception distortion trade-off~\cite{blau2018perception}. 

%The rest of the paper is organized as follows. The existing techniques for ultrasound image compression are briefly described in Sec.~\ref{prior art}. The details of our proposed framework is elucidated in Sec.~\ref{method}. The description of the data used in the experiments, results and discussion of the findings are detailed in Sec.~\ref{results} and the work is concluded in Sec.~\ref{conclusion}.

\section{Prior Art}
\label{prior art}

The significant growth in digital imaging over the last four decades has resulted in an increase in complexity associated with transmission and storage of images. International Standards Organization (ISO), and International Electro-technical Commission (IEC) jointly introduced the first standard for image compression popularly also referred to as the Joint Photographic Experts Group (JPEG)~\cite{pennebaker1992jpeg}. These developments were later on followed by genesis of new standards like the JPEG2000~\cite{boliek2002jpeg}, JPEG-LS~\cite{weinberger2000loco}, JPEG-XR~\cite{dufaux2009jpeg} and HEVC/H.265~\cite{ohm2012comparison} for video compression.

In line with the growth of digitally acquired and stored medical images, the requirement of standards for digital imaging and communication in medicine (DICOM)~\footnote{\texttt{ftp://medical.nema.org/medical/dicom/1992-1995/}} also grew. Subsequently, JPEG, JPEG 2000, JPEG-LS, H.264, and HEVC~\footnote{\texttt{ftp://medical.nema.org/medical/dicom/supps/PC/}} were included within DICOM and used for radiological images, and information on clinically acceptable compression factors was laid out for each of the specific imaging modalities by different societies~\cite{liu2017current}. 

However, use of compression in US images has been limited to JPEG and JPEG2000~\cite{liu2017current} and restricted for use in anatomical regions like the breast, musculoskeletal region and in pediatric imaging. Recently, methods employing deep neural networks have been introduced which allow high-density compression of medical images while retaining diagnostically relevant features~\cite{kar2018fully}. The motivation of this paper is to address this challenge by introducing a framework with the primary ability to preserve key information rendered through speckles to enable its clinically relevant use.

\section{Methodology}
\label{method}

The image compression framework we propose here consists of a compressor and a decompressor block, with the compressed file constituting the commonly exchanged information between them. The key idea is to segment an image into anatomically relevant regions and transmit it to the decompressor which is essentially an image generator ~\cite{tom2018simulating}, that generates the image from the compressed file.

\subsection{Compressor}
\label{comp}
The compressor consists of a segmentation engine which splits the ultrasound image into anatomical and pathological segments and transmits them. We demonstrate generalizability by implementing the segmentation using two different methods which have been recently published, (i) a classical machine learning based approach with cross frame belief propagating iterative random walks~\cite{china2018anatomical} and, (ii) a convolutional neural network based approach implemented with U-Net~\cite{iglovikov2018ternausnet}.

\textbf{Classical machine learning based approach for segmentation:} This approach~\cite{china2018anatomical} is based on the statistical mechanical understanding of ultrasound-tissue interaction. A parametric model of ultrasonic speckle statistics is estimated and used to learn a random forest classifier for pixelwise classification of tissues. This model is used for contour initialization. Subsequently, iterative random walks are used for correcting the contours. Gradient vector flow based belief propagation is then applied to subsequent neighboring frames for initializing the random walks, and this process is performed iteratively for volume segmentation. 

\textbf{Convolutional neural network approach for segmentation:} A semantic segmentation approach based on U-Net~\cite{iglovikov2018ternausnet} is also used, where initialization is performed with VGG11~\cite{simonyan2014very} weights in the encoder unit. The decoder weights are randomly initialized. The network generates a pixel-wise segmentation response map. The ReLU activation function is used throughout the network following convolution layers. In the final layer of the decoder, the sigmoid activation function is used to generate a pixel-wise segmentation response map. The weighted cross-entropy (WCE) loss function is used, giving higher weights to the pixels closer to the boundary using using the morphological distance transform.

\textbf{File format:} The file format for transmission of compressed data contains four necessary pieces of information. Chain codes of the contour for different classes of tissue are stored in the probe geometry specific polar coordinate format. A number of scan lines that make up the polar image is included for reconstruction of the segmented contours in the decompressor. The frame size in the Cartesian coordinate system is also included. The acquisition frequency is included as well, for the reconstruction of the images using a generative model.
\subsection{Decompressor}

In this section, an adversarially trained generative convolutional network is employed to simulate patho-realistic ultrasound images from tissue echogenicity maps recreated from the received file. This generator involves two stages and is based on the framework proposed in ~\cite{tom2018simulating}.

\textbf{Stage 0: Physics based simulation:} This first stage simulation is performed using a pseudo B-mode physics based simulator %~\footnote{\texttt{https://in.mathworks.com/matlabcentral\\/fileexchange/34199}}
which works on the principle of linear and space invariant nature of point spread function (PSF) of speckle in the image~\cite{bamber1980ultrasonic}.

\textbf{Stage I: Speckle intensity and Point spread function learning:}
In this stage, we modify the architecture in ~\cite{tom2018simulating} to enable direct learning of the mapping from stage 0 simulated results to patho realistic ultrasound images, using a single generator. This allows for better preservation of the ground truth annotation which is an important diagnostically relevant feature. The proposed single stage model has the added advantage of being more computationally efficient, reducing the inference time by half. 

The generator has an encoder-decoder architecture that enables it to learn the intensity mapping and the point spread function. The feature maps are fed to residual blocks, by nearest neighbour upsampling followed by convolutions. The discriminator has downsampling blocks that brings down the dimension to $4\times4$ followed by a $1\times1$ convolution layer and a fully connected layer for prediction. The LeakyReLU activation function is used in the downsampling blocks. Similar to the network in \cite{tom2018simulating} a self regularization term is also included in the GAN loss for preserving the ground truth information, whose value was chosen as 0.01 via experimentation.

\begin{figure*}[t]
\centering
\subfigure[Original Image]{\label{org} {\setlength\fboxsep{0pt}\setlength\fboxrule{0pt}\fbox{\includegraphics[width=0.15\textwidth]{Original.png}}}}
  \subfigure[UltraCompression]{\label{Ultracompression} {\setlength\fboxsep{0pt}\setlength\fboxrule{0pt}\fbox{\includegraphics[width=0.15\textwidth]{Ultracompression.png}}}}
 \subfigure[JPEG]{\label{jpeg} {\setlength\fboxsep{0pt}\setlength\fboxrule{0pt}\fbox{\includegraphics[width=0.15\textwidth]{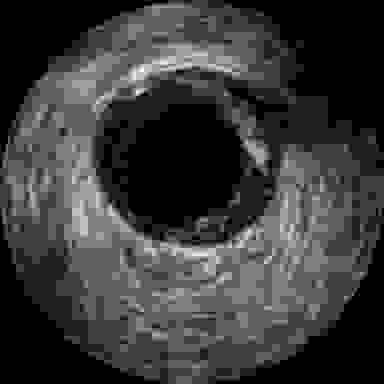}}}}
  \subfigure[JPEG2000]{\label{jpeg2000} {\setlength\fboxsep{0pt}\setlength\fboxrule{0pt}\fbox{\includegraphics[width=0.15\textwidth]{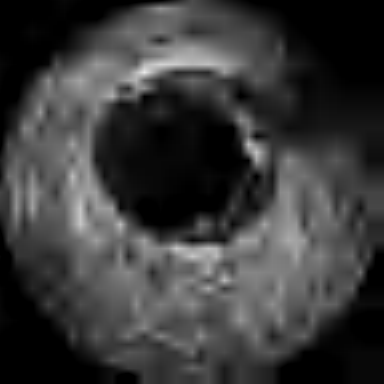}}}}
\subfigure[BPG]{\label{bpg} {\setlength\fboxsep{0pt}\setlength\fboxrule{0pt}\fbox{\includegraphics[width=0.15\textwidth]{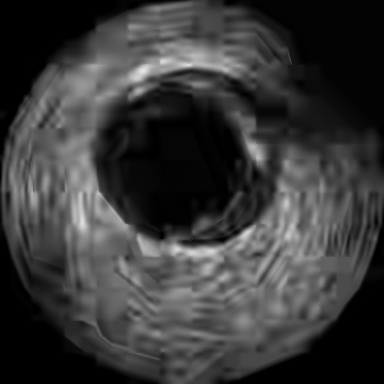}}}}
\subfigure[WebP]{\label{webp} {\setlength\fboxsep{0pt}\setlength\fboxrule{0pt}\fbox{\includegraphics[width=0.15\textwidth]{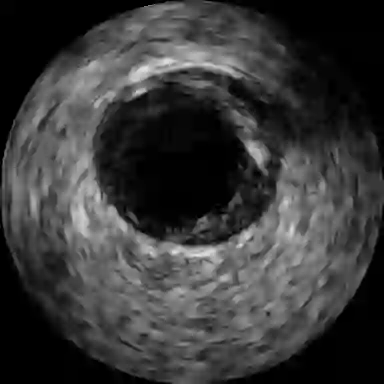}}}}
\caption{Visualization of decompressed images.}
\label{Vis}
\end{figure*}

\section{Experiments, Results and Discussion}
\label{results}

In this experiment, we have used the IVUS pullback data from the border detection in IVUS Challenge~\footnote{\texttt{http://www.cvc.uab.es/IVUSchallenge2011/dataset.html}}is used. The dataset consists of $10$ pullbacks with one pullback per patient, acquired at $20$ MHz~\cite{balocco2014standardized}. One pullback was used for reporting performance and remaining nine were used for training. Augmentation by axially rotating the pullbacks by 30\degree over 12 steps was done. Implementation has been done on Ubuntu $16.04$ LTS OS, Python 3.6, PyTorch 0.5, CUDA $9.2$ and cuDNN $6.1$ for acceleration with a Nvidia Tesla K40c GPU with 12GB of DDR4 RAM on a PC with Intel Core i5-8600K processor and 32 GB of system RAM.

The compressor with classical machine learning approach was implemented following \cite{china2018anatomical}. The UNet~\cite{iglovikov2018ternausnet} based compressor was trained with learning rate $1 \times 10^{-3}$ and batch size 14 using the Adam optimizer. The decompressor was trained with an initial learning rate of $2 \times 10^{-4}$ and a weight decay of $0.5$ per $100$ epochs over $1,200$ epochs with a batch size of $64$, following ~\cite{tom2018simulating}.

\begin{table}[h]
\centering
\caption{Inter-tissue Jensen Shannon (JS) divergence. Higher value indicates better contrast between tissue pairs. Also refer Fig.~\ref{Perception}. Ext. denotes external elastic luminae. Numbers in parentheses correspond to standard deviation.}
\small
\label{interJS}
\begin{tabular}{p{2.2cm}p{1.5cm}p{1.5cm}p{1.5cm}}
\hline
\hline
\textbf{} & \textbf{Lumen vs. Media}   & \textbf{Media vs. Ext.} & \textbf{Lumen vs. Ext.} \\ \hline 
{Original}  & {$0.18(0.05)$} & { $0.19(0.04)$ } & { $0.11(0.08)$ } \\ 
{BPG}  & { $0.17(0.10)$ } & { $0.14(0.06)$ } & { $0.11(0.10)$ } \\ 
{JPEG}  & { $0.11(0.07)$ } & { $0.09(0.07)$ } & { $0.14(0.10)$ } \\ 
{JPEG2000}  & { $0.13(0.10)$ } & { $0.16(0.05)$ } & { $0.11(0.09)$ } \\ 
{WebP}  & { $0.18(0.06)$ } & { $0.16(0.09)$ } & { $0.13(0.11)$ } \\ 
 \textbf{UltraCompression}  & { $0.21(0.09)$ } & { $0.18(0.07)$ } & { $0.18(0.10)$ } \\
        \hline
        \hline
    \end{tabular}
\end{table}

\begin{table}[h]
\centering
\caption{Intra-tissue JS divergence. Lower value indicates better preservation of speckle statistics.}
\small
\label{intraJS}
\begin{tabular}{p{2.5cm}p{1.2cm}p{1.2cm}cp{1.2cm}p{1.2cm}p{1cm}}
\hline
\hline
\textbf{Method}  &\textbf{Lumen}   & \textbf{Media} & \textbf{Ext.} \\ \hline 
BPG & $0.16(0.11)$  &  $0.14(0.08)$ &  $0.21(0.16)$  \\ 
JPEG & $0.23(0.14)$  &  $0.20(0.09)$ &  $0.29(0.16)$  \\ 
JPEG2000 & $0.21(0.13)$  &  $0.17(0.11)$ &  $0.22(0.10)$  \\ 
WebP & $0.12(0.11)$  &  $0.10(0.08)$ &  $0.15(0.09)$  \\ 
\textbf{UltraCompression} & $0.10(0.07)$  &  $0.09(0.06)$ &  $0.12(0.04)$  \\ 
        \hline
        \hline
    \end{tabular}
\end{table}

\begin{table}[h]
\centering
\caption{Intra-tissue JS divergence in attenuation map. Lower value indicates similarity in estimated signal attenuation.}
\small
\label{intraJSAttenuation}
\begin{tabular}{p{2.5cm}p{1.2cm}p{1.2cm}cp{1.2cm}p{1.2cm}p{1cm}}
\hline
\hline
\textbf{Method}  &\textbf{Lumen}   & \textbf{Media} & \textbf{Ext.} \\ \hline 
BPG & $0.14(0.07)$  &  $0.12(0.10)$ &  $0.15(0.10)$  \\ 
JPEG & $0.24(0.15)$  &  $0.19(0.09)$ &  $0.30(0.14)$  \\ 
JPEG2000 & $0.14(0.09)$  &  $0.13(0.04)$ &  $0.16(0.09)$  \\ 
WebP & $0.10(0.10)$  &  $0.10(0.06)$ &  $0.13(0.07)$  \\ 
\textbf{UltraCompression} & $0.07(0.03)$  &  $0.08(0.05)$ &  $0.08(0.08)$  \\ 
        \hline
        \hline
    \end{tabular}
\end{table}

The original and the decompressed images are compared using the inter-tissue Jensen-Shannon (JS) divergence of their speckle appearance and the results are reported in Table~\ref{interJS}. We compared the proposed method with the prior art using intra-tissue JS divergence assessed between speckle statistics in Table~\ref{intraJS} and divergence in attenuation in Table~\ref{intraJSAttenuation}. Subsequently, we compared the impact on image segmentation using two strategies for segmentation. In the first case, we used a CNN within the compressor and a classical machine learning based approach for segmentation during validation, whose results are reported in Table~\ref{classsical}. In the second case, the compression and segmentation approaches were reversed and the results are reported in Table~\ref{CNN}.

\begin{table}[h]
\centering
\caption{Evaluating segmentation performance on decompressed images using \cite{china2018anatomical}. SE denotes sensitivity, SP denotes specificity, PPV denotes Positive Predictive Value.}
\small
\label{classsical}
\begin{tabular}{p{1.3cm}p{1.3cm}p{1.3cm}p{1.3cm}p{1.3cm}}
\hline
\hline
\textbf{Method} & \textbf{SE}  &\textbf{SP}   & \textbf{Dice} & \textbf{PPV} \\ \hline 
BPG & $0.97(0.02)$  &  $0.66(0.12)$ &  $0.65(0.12)$ &  $0.66(0.14)$ \\ 
JPEG & $0.87(0.08)$  &  $0.88(0.11)$ &  $0.61(0.24)$ &  $0.54(0.25)$ \\ 
JPEG2000 & $0.98(0.01)$ &  $0.61(0.14)$ &  $0.62(0.08)$  & $0.65(0.09)$\\ 
WebP & $0.98(0.01)$  &  $0.81(0.12)$ &  $0.75(0.09)$  &  $0.72(0.11)$\\ 
\textbf{UltraComp.} & $0.98(0.02)$  &  $0.85(0.14)$ &  $0.87(0.12)$  &  $0.90(0.09)$\\ 
        \hline
        \hline
\end{tabular}
\end{table}

\begin{table}[h]
\centering
\caption{Evaluating segmentation performance on decompressed images using \cite{iglovikov2018ternausnet}.}
\small
\label{CNN}
\begin{tabular}{p{1.3cm}p{1.3cm}p{1.3cm}p{1.3cm}p{1.3cm}}
\hline
\hline
\textbf{Method} & \textbf{SE}  &\textbf{SP}   & \textbf{Dice} & \textbf{PPV} \\ \hline 
BPG & $0.98(0.02)$  &  $0.58(0.35)$ &  $0.63(0.30)$ &  $0.82(0.14)$ \\ 
JPEG & $0.96(0.02)$  &  $0.51(0.33)$ &  $0.56(0.27)$ &  $0.72(0.18)$ \\ 
JPEG2000 & $0.98(0.01)$ &  $0.58(0.15)$ &  $0.55(0.18)$  & $0.59(0.26)$\\ 
WebP & $0.98(0.02)$  &  $0.67(0.35)$ &  $0.70(0.30)$  &  $0.82(0.13)$\\ 
\textbf{UltraComp.} & $0.98(0.02)$  &  $0.87(0.11)$ &  $0.89(0.11)$  &  $0.92(0.06)$\\ 
        \hline
        \hline
\end{tabular}
\end{table}

Here, an uncompressed polar image occupies $384 \times 256 \times 8$ bits. For the chain code used, the starting point occupies $1 \times 2 \times 16$ bits and the remaining points occupy $255 \times 2$ bits. Two contours for lumen and media are stored and thus the compression factor becomes $(384 \times 256 \times 8)/(((1 \times 2 \times 16) + (255 \times 2)) \times 2) \approx 725$. In spite of achieving a high compression ratio, the decompressed images have very low intra-tissue JS divergence values, indicating a high degree of similarity to the original images. From Table~\ref{interJS}, it is observed that the tissue-specific layer mapping is better than in the existing methods because the inter-tissue JS divergence is higher and nearly similar to that of the original images. Additionally, inter-tissue statistical mechanics has been mapped properly in the decompressed image since the classical method based on the statistical mechanical understanding of ultrasound-tissue interaction has been able to segment the lumen and external elastic luminae precisely. Through a paired visual Turing test, it was observed that there was a 50\% chance of identifying the real from the decompressed images ~\cite{tom2018simulating}. Also, the compressor and decompressor blocks in this framework can be modified with other segmentation and generation methods to further improve performance. %During inference, for compression, JPEG (implemented on OpenJPEG library) takes 0.021 seconds whereas the proposed method takes 0.068 seconds. During decompression, JPEG takes 0.017 seconds whereas the proposed method takes 0.100 seconds. The validation was performed on a CPU platform consisting of $2 \times$ Intel Xeon 8160 CPU with $12 \times 32$GB DDR4 ECC Regd. RAM. 

\section{Conclusion}
\label{conclusion}

In this paper, we have proposed a framework for high-density compression of ultrasound volumes. This framework involves two parts, a compressor with a segmentation block and a decompressor with a generation block. Both the segmentation and generation blocks can be customized with different algorithms. Here we we have used classical machine learning and convolutional neural network alternatively in the compressor block, and a generative adversarial network in the decompressor block. The quality of image compression has been evaluated using inter- and intra-tissue JS divergence between the original and the decompressed images. 

\bibliographystyle{IEEEbib}
\small
\bibliography{refs}

\end{document}